\documentclass[letterpaper]{article} 
\usepackage[]{aaai23}  
\usepackage{times}  
\usepackage{helvet}  
\usepackage{courier}  
\usepackage[hyphens]{url}  
\usepackage{graphicx} 
\usepackage{natbib}  
\usepackage{caption} 
\usepackage{algorithm}
\usepackage{algpseudocode}
\usepackage[caption=false,font=footnotesize]{subfig}
\usepackage{eqparbox}
\usepackage{xcolor}
\newdimen{\algindent}
\algnewcommand\LeftComment[2]{%
\hspace{#1\algindent}$\triangleright$ \eqparbox{COMMENT}{#2} \hfill %
}

\usepackage{amsmath}
\usepackage{amssymb}
\usepackage{array,multirow}
\graphicspath{{figures/}}

\usepackage{newfloat}
\usepackage{listings}
\DeclareCaptionStyle{ruled}{labelfont=normalfont,labelsep=colon,strut=off} 
\floatstyle{ruled}
\newfloat{listing}{tb}{lst}{}
\floatname{listing}{Listing}
\usepackage{xspace}
\newcommand{\trackername}{Minkowski Tracker\xspace}
\newcommand{\networkname}{Minkowski Tracker R-CNN\xspace}

\title{\trackername: A Sparse Spatio-Temporal R-CNN \\for Joint Object Detection and Tracking}

\author {
    JunYoung Gwak,
    Silvio Savarese,
    Jeannette Bohg
}
\affiliations {
    Stanford University\\
    jgwak@cs.stanford.edu, ssilvio@stanford.edu, bohg@stanford.edu
}

\begin{document}

\maketitle

\begin{abstract}

Recent research in multi-task learning reveals the benefit of solving related problems in a single neural network. 3D object detection and multi-object tracking (MOT) are two heavily intertwined problems predicting and associating an object instance location across time. However, most previous works in 3D MOT treat the detector as a preceding separated pipeline, disjointly taking the output of the detector as an input to the tracker. In this work, we present \trackername, a sparse spatio-temporal R-CNN that jointly solves object detection and tracking. Inspired by region-based CNN (R-CNN), we propose to solve tracking as a second stage of the object detector R-CNN that predicts assignment probability to tracks. First, \trackername takes 4D point clouds as input to generate a spatio-temporal Bird's-eye-view (BEV) feature map through a 4D sparse convolutional encoder network. Then, our proposed \texttt{TrackAlign} aggregates the track region-of-interest (ROI) features from the BEV features. Finally, \trackername updates the track and its confidence score based on the detection-to-track match probability predicted from the ROI features. We show in large-scale experiments that the overall performance gain of our method is due to four factors: 1. The temporal reasoning of the 4D encoder improves the detection performance 2. The multi-task learning of object detection and MOT jointly enhances each other 3. The detection-to-track match score learns implicit motion model to enhance track assignment 4. The detection-to-track match score improves the quality of the track confidence score. As a result, \trackername achieved the state-of-the-art performance on Nuscenes dataset tracking task without hand-designed motion models.

\end{abstract}

\section{Introduction}

3D Multi-object tracking (MOT) allows autonomous agents to perceive the motion of various entities in their surroundings. MOT is one of the core perception problems of various real-time robotics applications such as autonomous driving~\cite{luo2021exploring,petrovskaya2008model} and collaborative robotics~\cite{gross2011ll,erol2018improved}. With the increasing demand for AI-based, real-world automation~\cite{maurer2016autonomous} and better affordability of 3D sensors~\cite{hecht2018lidar}, the need for robust 3D MOT has become ever higher. Consequently, the advent of multiple large-scale 3D MOT dataset~\cite{geiger2013vision, caesar2020nuscenes, martin2021jrdb, sun2020scalability} and improved techniques to process 3D data~\cite{graham2014spatially, choy20194d} have led to a recent significant advances in 3D MOT.

The most popular multi-object tracking (MOT) framework solves correspondences among detections over time, commonly known as a tracking-by-detection paradigm. Most previous approaches to tracking-by-detection treat detection and tracking as a separate module, having an object detector that runs frame-by-frame and then establishing correspondences over time. This pipelined approach is so widely adopted that many MOT challenges~\cite{MOTChallenge2015, MOT19_CVPR} provide baseline public object detection results, and most trackers in the Nuscenes challenge~\cite{caesar2020nuscenes} evaluate using the same object detection method~\cite{centerpoint} for a fair comparison. However, recent research in multi-task learning reveals that jointly solving related tasks improves the generalization performance of all the tasks~\cite{zhang2021survey}, which could extend to detection and tracking.

The last decade of machine learning research has revealed two important lessons: 1. End-to-end learning tends to achieve better results than pipelined approaches. 2. Related learning tasks can help each other, as demonstrated in transfer learning~\cite{zhuang2020comprehensive}, multi-task learning~\cite{zhang2021survey}, and cross-task learning~\cite{zamir2020robust}. The most common tracking-by-detection approach builds hand-designed filters outside of the neural network, not being end-to-end and sometimes slow. Furthermore, detection and tracking are related problems predicting the object instance and location over time that could potentially benefit from multi-task learning. To this end, we propose \trackername, a sparse spatio-temporal R-CNN for joint object detection and tracking.

The input space of our 3D MOT is in 4D Minkowski spacetime, a combination of 3D Euclidean space and time into a 4D manifold. Being a true end-to-end system, \trackername takes a sequence of 3D point clouds of length $\mathcal{T}$ as input and directly processes it using a 4D sparse encoder. The explicit temporal reasoning of the 4D encoder is robust to temporal changes such as occlusion, truncation, and transformation. Furthermore, the 4D encoder generates a spatio-temporal bird's-eye-view (BEV) feature map, which enables our tracking framework.

Inspired by the success of region-based CNN (R-CNN) in predicting additional information regarding detected instances such as mask~\cite{he2017mask} and 3D mesh~\cite{gkioxari2019mesh}, \trackername solves tracking-by-detection MOT by predicting the match probability of each detected instance to current tracks. To enable matching between detections and tracks, our proposed \texttt{TrackAlign} samples collective region-of-interest (ROI) features from the spatio-temporal BEV feature map. As a result, \trackername learns implicit motion models by predicting the detection's match probability to the tracks. Moreover, the multi-task learning of object detection and MOT jointly enhances the performance of each other, as we will show in experiments.

Lastly, prediction confidence is crucial information for any system that takes uncertain inputs. For example, informative object track confidence could be helpful for robotics applications such as anomaly detection and decision making. Consequently, the recent MOT evaluation metric AMOTA~\cite{amota} takes track confidence into account. A common practice of previous works is to take an average of the object detection confidences within the track, which lacks information on the confidence of associations across the track. \trackername complements this missing information using the network-predicted detection-to-track match probability scores.

To summarize, we propose \trackername, a sparse spatio-temporal R-CNN for joint object detection and tracking. Our proposed R-CNN structure solves tracking by sharing features with the detector to learn implicit motion models. Additionally, we propose \texttt{TrackAlign}, a spatio-temporal region of interest (ROI) feature aggregation function to sample collective ROI features between detections and tracks. The performance gain of our method, as we ablate in Table~\ref{table:ablation}, is due to the four factors:

\begin{itemize}

    \item \trackername improves 3D object detection and tracking by reasoning directly on 4D spatio-temporal space using a 4D sparse convolutional encoder.
    
    \item The multi-task learning of 3D object detection and multi-object tracking in the same network enhances the performance of each other.

    \item \trackername learns implicit motion models as the second stage of a region-based CNN (R-CNN) by predicting the detection's match probability to the tracks, enhancing the quality of track assignment.

    \item \trackername offers more robust track confidence by incorporating the detection-to-track match probability score.

\end{itemize}
\section{Related Work}

\subsection{LiDAR-based 3D object detection}

Most modern 3D object detectors for outdoor autonomous driving scenarios follow a similar structure. First, a point cloud encoder encodes 3D point clouds into a BEV feature. Second, tracking heads infer 3D object detections on the extracted BEV features.

First, for the point cloud encoder, VoxelNet~\cite{zhou2018voxelnet} proposes to voxelize point clouds using a 3D voxel feature encoder based on PointNet~\cite{qi2017pointnet} followed by a 3D CNN. PointPillar~\cite{lang2019pointpillars} proposes to encode 3D point clouds into a 2D pillar feature map followed by a 2D CNN. SECOND~\cite{yan2018second} makes VoxelNet efficient by using sparse convolution~\cite{graham2014spatially} instead of a dense 3D CNN. The point cloud encoder of \trackername is a 4D sparse convolutional encoder. The 4D encoder improves object detection by reasoning over a long temporal horizon, being robust to temporal changes such as occlusion, truncation, and transformation. Additionally, the 4D encoder outputs a spatio-temporal BEV feature map to enable our tracking framework.

Second, the tracking heads infer 3D object detections on the BEV feature map. Point R-CNN~\cite{shi2019pointrcnn} and Voxel R-CNN~\cite{deng2021voxel} proposed two-stage region proposal networks similar to Faster R-CNN~\cite{ren2015faster}. CenterPoint~\cite{centerpoint} proposes a two-stage anchor-free keypoint detector network. 3DSSD~\cite{yang20203dssd} proposes a anchor-free single stage 3D detection network. Our proposed method is agnostic to the choice of object detection heads. We use CenterPoint as our baseline detector network for a fair comparison against previous works in MOT.

\subsection{Region-based CNN}

Region-based CNN (R-CNN) is a neural network architecture that infers scene knowledge based on region-of-interest (ROI) features. The initial goal of R-CNN was object detection~\cite{girshick2015fast,ren2015faster}, extracting ROI features based on region proposals to predict and refine object detection parameters. As the technique matured, the ROI features found a new use for predicting additional information about the detected objects. Mask R-CNN~\cite{he2017mask} proposes to solve instance segmentation as the second stage of object detection by making an object mask prediction from bilinear-interpolated ROI features. Mesh R-CNN~\cite{gkioxari2019mesh} proposes reconstructing a corresponding 3D mesh from the detection's ROI features. The proposed \trackername similarly extends object detection networks to multi-object tracking using a spatio-temporal ROI feature. Our proposed \texttt{TrackAlign} samples collective ROI features of detections and tracks to predict the detected object's match likelihood to current tracks.

\subsection{LiDAR-based 3D multi-object tracking}


There are numerous track association algorithms of detection-to-track trackers. The most simple yet powerful association algorithm is the global nearest neighbor (GNN) match~\cite{rana2014simpletrack}, which associates a detection with its closest track on the Bird's-Eye View (BEV) global Euclidean coordinate space. CenterPoint~\cite{centerpoint} slightly improves this algorithm by backprojecting a detection based on its predicted velocity into the timestep of the track and then performs GNN Matching there. Some use Kalman filters~\cite{kalman1960new} to find a posterior estimate of a tracked object instance. A simple motion model predicts a tracked object's next state which is then matched to a detection based on the Mahalanobis distance~\cite{chiu2020probabilistic} or IoU/GIoU~\cite{rezatofighi2019generalized, Weng2020_AB3DMOT_eccvw}. Others use random finite sets (RFS)~\cite{pang20213d, liu2022gnn}, factor graphs~\cite{poschmann2020factor, liang2022neural}, and graph neural networks~\cite{zaech2022learnable, braso2020learning, weng2020gnn3dmot} for track association and management. Unlike previous works, our network directly predicts associations between the detections and tracks, learning an implicit motion model.

In addition to the association algorithms, \cite{chiu2020probabilistic, weng2020gnn3dmot, baser2019fantrack} propose to fuse the aforementioned motion models with appearance features to improve track management quality. Additionally, \cite{wang2021immortal,wu20213d} propose maintaining invisible tracks to prevent ID switches during occlusions. The contribution of \trackername is complementary to these efforts, and it may benefit from multi-modality input and advanced track management algorithms.

\section{Method}

\begin{figure*}[t]
    \centering
    \includegraphics[width=0.9\textwidth]{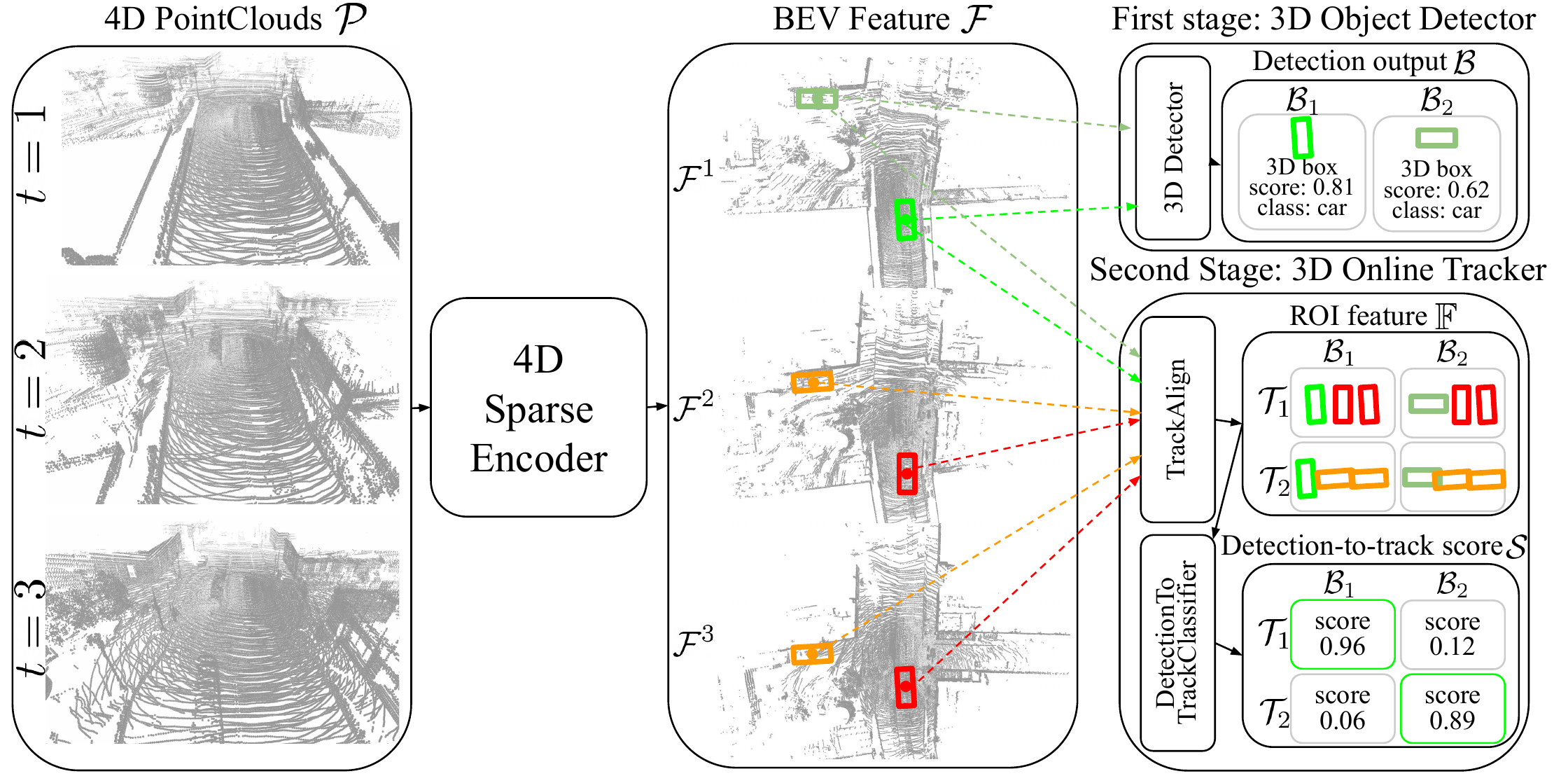}
    \caption{
        Visualization of \networkname with number of frames $T=3$. First, the 4D sparse encoder takes 4D point clouds $\mathcal{P}$ as an input to generate a spatio-temporal Bird's-eye-view feature map $\mathcal{F}$. Then, the 3D object detector outputs 3D bounding boxes $\mathcal{B}$. The next stage is an online tracker, which first extracts track ROI features $\mathbb{F}$ using our proposed \texttt{TrackAlign}. Finally, the \texttt{DetectionToTrackClassifier} predicts track match probability $\mathcal{S}$ for each detection.
    }
    \label{fig:overview}
\end{figure*}

In this section, we illustrate the \trackername in detail. First off, \trackername deals with spatio-temporal data. For consistent notation of time, let $T$ be the maximum number of frames. Additionally, let $t$ and any superscript $\cdot^t$ be a relative temporal index, where $t=1$ is the current and the latest frame where we detect objects, $t=2$ is a previous frame, and $t=T$ is the oldest frame the network processes.

The outline of our method is in Algorithm~\ref{alg:overview}. The input to \trackername is a 4D spatio-temporal point cloud $\mathcal{P}$, which at each timestamp is a temporal stack of up to $T$ previous 3D lidar point cloud observations.

The core component of our framework is \networkname, a two-stage detection-to-track sparse spatio-temporal R-CNN, visualized in Figure~\ref{fig:overview}. First, the 4D sparse encoder takes 4D point clouds $\mathcal{P}$ as input to generate a spatio-temporal Bird's-eye-view feature map $\mathcal{F}$ (Line~\ref{mt:alg1:encoder}). Then, the 3D object detector head outputs 3D detection bounding boxes $\mathcal{B}$ on the BEV feature $\mathcal{F}^1$ at the current frame $(t=1)$ (Line~\ref{mt:alg1:detector}). The next stage is an online tracker, which first extracts spatio-temporal ROI features $\mathbb{F}$ from a combination of detections $\mathcal{B}$ and current tracks $\mathcal{T}$ using our proposed \texttt{TrackAlign} (Line~\ref{mt:alg1:trackalign}). Then, the \texttt{DetectionToTrackClassifier} outputs a match probability score $\mathcal{S}$ between the detections and tracks from $\mathbb{F}$ (Line~\ref{mt:alg1:classifier}).

Finally, given the network output detections $\mathcal{B}$ and detection-to-track scores $\mathcal{S}$, \trackername updates the tracks $\mathcal{T}$ with a simple Hungarian matching over $\mathcal{S}$ and calculates its track confidence score (Line~\ref{mt:alg1:conf}). In the following subsections, we detail each component of our framework.

\begin{algorithm}
\caption{Outline of \trackername}\label{alg:overview}
\begin{algorithmic}[1]
\State\textbf{Input} $\{\mathcal{P}^t\}\forall t$: 3D lidar point clouds at relative time $t$
\State\textbf{Input} $T$: Maximum temporal axis size
\State\textbf{Output} $\mathcal{T}$: Tracks
\State$\mathcal{T}\gets \{\}$
\For{$i\in\{1, 2, \ldots\}$}
    \State$\mathcal{P}\gets \{\mathcal{P}^t\}\forall t \in [1, \min(T, i)]$
    \Procedure{\networkname}{$\mathcal{P}, \mathcal{T}$}
    \State\textbf{Input} $\mathcal{P}$: 4D spatio-temporal lidar point clouds
    \State\textbf{Input} $\mathcal{T}$: Tracks
    \State\textbf{Output} $\mathcal{B}$: Detection bounding boxes
    \State\textbf{Output} $\mathcal{S}$: Detection-to-track match probability
    \State\LeftComment{0}{First stage: 3D object detector}
    \State $\mathcal{F}\gets\texttt{4DSparseEncoder}(\mathcal{P})$ \label{mt:alg1:encoder}
    \State$\mathcal{B}\gets\texttt{3DObjectDetector}(\mathcal{F}^1)$ \label{mt:alg1:detector}
    \State\LeftComment{0}{Second stage: 3D online tracker}
    \State$\mathbb{F}\gets\texttt{TrackAlign}(\mathcal{F}, \mathcal{B}, \mathcal{T})$ \label{mt:alg1:trackalign}
    \State$\mathcal{S}\gets\texttt{DetectionToTrackClassifier}(\mathbb{F})$ \label{mt:alg1:classifier}
    \EndProcedure
    \State\LeftComment{0}{Track match and track confidence update}
    \State$\mathcal{T} \gets \texttt{TrackManager}(\mathcal{B}, \mathcal{T}, \mathcal{S})$ \label{mt:alg1:conf}
\EndFor
\end{algorithmic}
\end{algorithm}

\subsection{4D sparse encoder}

\networkname takes 4D spatio-temporal sparse point cloud in Minkowski timespace as an input. $\mathcal{P}=\{(x, y, z, t, r)_i\}$ where $(x, y, z)$ is 3D location in Euclidean space, $t$ is time in a relative temporal index, and $r$ is reflectance measurement. The output of the encoder is a spatio-temporal Bird's-eye-view (BEV) feature map $\mathcal{F}=\{\mathcal{F}^1, \mathcal{F}^2, \ldots, \mathcal{F}^T\}$.

In short, our encoder is a 4D-equivalent of SECOND~\cite{yan2018second} with some adjustments on the temporal axis to maintain its temporal shape. First, the 4D voxelizer quantizes the points in 3D Euclidean space while retaining the temporal index. Let the 3D location of the point clouds range within $W$, $H$, $D$ along the $X$, $Y$, $Z$ axis respectively. The voxelizer groups points along the 3D Euclidean voxel grid of size $v_W$, $v_H$, $v_D$ for each $t$. Then, the voxel feature encoding (VFE) layer~\cite{zhou2018voxelnet} extracts voxel features for each voxel group. Note that the voxelizer does not apply any grouping nor VFE along the temporal axis to retain the temporal dimension of the voxel tensor. The resulting 5D sparse tensor is of size $(C_{in} \times T \times \lfloor \frac{D}{v_D} \rfloor \times \lfloor \frac{H}{v_H} \rfloor \times \lfloor \frac{W}{v_W} \rfloor)$ where $C_{in}$ is the size of the input feature channel.

Then, the 4D sparse convolutional middle encoder network processes the voxelized 5D sparse tensor to generate dense spatio-temporal BEV images $\mathcal{F}$. Our middle encoder follows the same architecture as that of SECOND~\cite{yan2018second} with the following modifications. First, our middle encoder uses 4D sparse convolution instead of 3D sparse convolution to enable spatio-temporal reasoning. Second, the network never strides along the temporal axis to preserve the temporal shape $T$. Given the encoder network spatial stride factor of st, the resulting dense 4D spatio-temporal BEV tensor is of size $\mathcal{F}=(C_{out} \times T \times \lfloor \lfloor \frac{H}{v_H} \rfloor \frac{1}{\text{st}} \rfloor \times \lfloor \lfloor \frac{W}{v_W} \rfloor \frac{1}{\text{st}}\rfloor)$ where $C_{out}$ is the output channel size and the $Z$ axis is squashed down by the network design. Finally, we split $\mathcal{F}$ along the temporal axis to get a 2D BEV feature images at each specific $t$, $\mathcal{F}=\{\mathcal{F}^1, \mathcal{F}^2, \ldots, \mathcal{F}^T\}$. As a result, our 4D sparse encoder fully propagates information across the entire 4D spatio-temporal space, explicitly reasoning on the 3D spatial domain and its temporal changes.

\subsection{3D object detector}

The first stage of \networkname, including the encoder above, is a 3D object detector. It takes a 2D BEV feature image $\mathcal{F}^1$ at the current frame $(t=1)$ as an input and outputs 3D bounding boxes $\mathcal{B}=\{(u, v, d, w, l, h, \alpha, vel, c, s_{\text{det}})\}$ of a center location $(u, v, d)$, 3D size $(w, l, h)$, rotation $\alpha$, velocity $vel$, semantic class $c$, and detection confidence score $s_{\text{det}}$ per box.

\networkname is agnostic to the choice of the detection algorithm. Without loss of generality, we chose CenterPoint~\cite{centerpoint} as our 3D detection model for fair comparison against most previous works. The performance of the tracker is heavily affected by that of the detector. \networkname is a modular network that can incorporate any research advance in the object detector to improve its general performance. For simplicity, we define any loss functions used to train the 3D object detector as $L_{\text{det}}$.

Although \networkname shares the identical detector head to CenterPoint~\cite{centerpoint}, there are two crucial differences that set our detection results apart. First, the explicit temporal reasoning of the 4D encoder is robust to temporal changes such as occlusion, truncation, and transformation. Second, the multi-task learning of object detection and tracking enhances the quality of each other. In our experiment, we will quantitatively ablate the two to demonstrate how \networkname improves detection quality without any changes in the detector head.

\begin{figure}[t]
    \centering
    \includegraphics[width=0.4\textwidth]{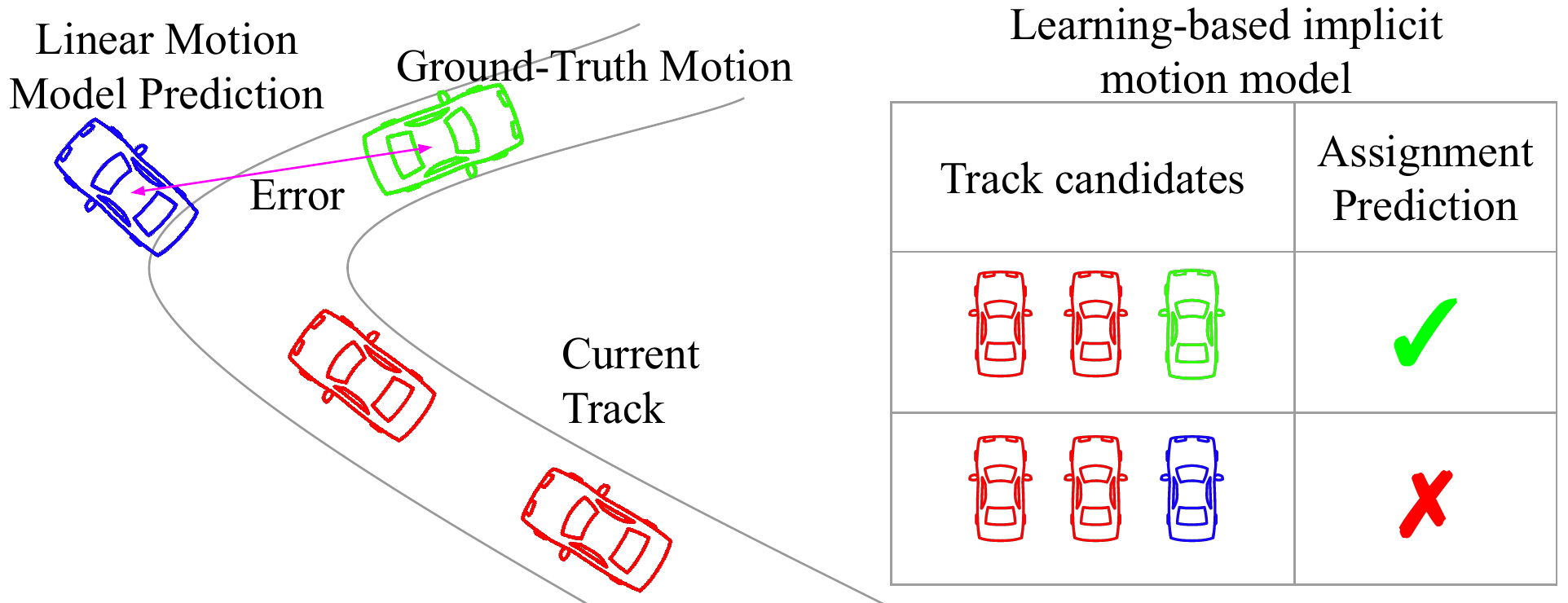}
    \caption{
        Visualization of a simple linear motion model compared to our proposed implicit motion model. \textcolor{red}{Red cars} represent current track, \textcolor{blue}{Blue car} is a prediction of a linear motion model, and \textcolor{green}{Green car} is a ground-truth motion. (left) A simple motion model may suffer from motions that are out of its model's scope. (right) In contrast, our proposed network-predicted implicit motion model can perceive the entire spatio-temporal scene to capture any motions.
    }
    \label{fig:motionmodel}
\end{figure}

\subsection{3D online tracker}

The second stage of our network is a 3D online tracker. The input to our tracker is:

\begin{itemize}

    \item 3D object bounding boxes from the detector $\mathcal{B}=\{\mathcal{B}_1, \mathcal{B}_2, \ldots, \mathcal{B}_N\}$ where $N$ is the number of detected objects.
    
    \item Current tracks $\mathcal{T}=\{\mathcal{T}_1, \mathcal{T}_2, \ldots, \mathcal{T}_M\}$ where $M$ is the number of tracks and each track $\mathcal{T}_i=\{\mathcal{T}_i^2, \mathcal{T}_i^3, \ldots, \mathcal{T}_i^T\}$ is a set of 3D bounding boxes of the same instance moving across the previous frame $t=2$ to the last frame $t=T$. Note that part of the track could be missing due to occlusion or missing detections (e.g. $\mathcal{T}_i=\{\mathcal{T}_i^2, \mathcal{T}_i^4, \mathcal{T}_i^5\}$).
    
    \item The spatio-temporal BEV image features $\mathcal{F}=\{\mathcal{F}^1, \mathcal{F}^2, \ldots, \mathcal{F}^T\}$ from the 4D encoder.

\end{itemize}

The tracker outputs a matrix $S=(N \times M)$ where each element is a match probability between the detections and tracks.

A common way to associate detections to tracks is based on a hand-designed filter that often uses simple linear motion models. The filters predict the next state of the track, and trackers measure the distance of the predicted states to the detections to build a cost matrix. However, simple motion models may not be sufficient to represent complicated real-world actions such as a curvy road as visualized in Figure~\ref{fig:motionmodel}. In contrast, a neural network can directly observe the car's motion along the road and associate its movement. Therefore, we propose to implicitly model motions within the neural network by directly predicting the match probability of detections to current tracks.

Furthermore, object detection and tracking are related problems, predicting the object instance and location over time. Research in multi-task learning reveals that solving related problems jointly with a shared network enhances each other~\cite{zhang2021survey}. However, most previous approaches to detection-to-track decouple detection and tracking as independent pipelines. To this end, we propose solving online tracking as the second stage of a detector by aligning spatio-temporal ROI features with our novel \texttt{TrackAlign}. Our experiment will quantitatively verify that matching tracks based on the learned detection-to-track matching score improves tracking quality.

\subsection{\texttt{TrackAlign}}

\begin{figure}[t]
    \centering
    \includegraphics[width=0.4\textwidth]{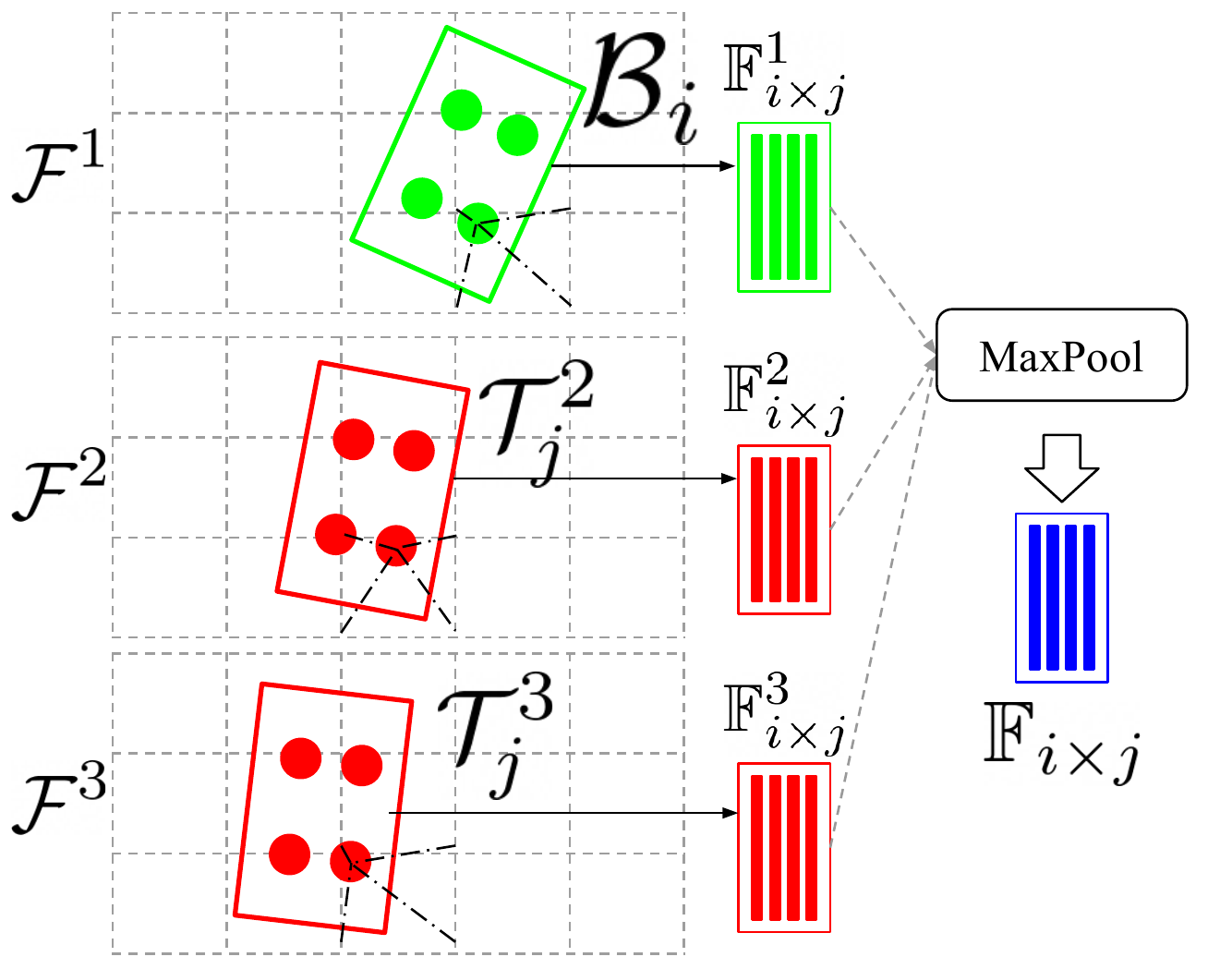}
    \caption{
        Visualization of \texttt{TrackAlign} with number of frames $T=3$, \textcolor{green}{$i$-th bounding box} $\mathcal{B}_i$, \textcolor{red}{$j$-th track} $\mathcal{T}_j=\{\mathcal{T}_j^2, \mathcal{T}_j^3\}$, and spatio-temporal BEV image features $\mathcal{F}=\{\mathcal{F}^1, \mathcal{F}^2, \mathcal{F}^3\}$. For each timestep, rotated \texttt{ROIAlign} samples bounding box features using bilinear interpolation. Then, we \texttt{\textcolor{blue}{MaxPool}} the features across the temporal axis.
    }
    \label{fig:TrackAlign}
\end{figure}

We visualize \texttt{TrackAlign} in Figure~\ref{fig:TrackAlign}. \texttt{TrackAlign} samples ROI features $\mathbb{F}=(N \times M)$ of a combination of $\mathcal{B}$ and $\mathcal{T}$. Without loss of generality, we describe how \texttt{TrackAlign} extracts the $i$th row and $j$th column of $\mathbb{F}$, $\mathbb{F}_{i \times j}$ from the $i$-th bounding box $\mathcal{B}_i$ and $j$-th track $\mathcal{T}_j$.

First, rotated ROIAlign~\cite{ren2015faster} extracts the temporal ROI feature $\mathbb{F}_{i \times j}^t$ of each timestep $t$ from a 2D BEV feature $\mathcal{F}^t$ and a 2D rotated bounding box $\mathbb{B}^t$. At the time of inference, $\mathbb{B}^1 = \texttt{BBox3Dto2D}(\mathcal{B}_i)$ and $\mathbb{B}^t = \texttt{BBox3Dto2D}(\mathcal{T}_j^t) \forall t \in [2, T]$. \texttt{BBox3Dto2D} is a function that converts 3D bounding box to a 2D BEV bounding box by removing its $Z$ axis elements, $d$ and $h$. In case tracks are missing at certain timestamps $t_{\text{miss}}$, we skip extracting the feature $\mathbb{F}_{i \times j}^{t_{\text{miss}}}=\emptyset$. Finally, we \texttt{MaxPool} all extracted ROI features across the temporal axis $\mathbb{F}_{i \times j}=\texttt{MaxPool}(\mathbb{F}_{i \times j}^1, \mathbb{F}_{i \times j}^2, \ldots, \mathbb{F}_{i \times j}^T)$.

There are three benefits of \texttt{TrackAlign} worth mentioning: 1. The aggregated feature $\mathbb{F}$ is aware of the heading of an instance since we use rotated \texttt{ROIAlign}. 2. The bilinear interpolation of \texttt{ROIAlign} allows sub-pixel reasoning. From a practical standpoint, the size of a pixel of the BEV image used in our network is 60cm which is bigger than the average human shoulder-to-shoulder distance, 40cm. 3. \texttt{TrackAlign} is agnostic to the length of the temporal window $T$ and missing detections in tracks, owing to the symmetric function \texttt{MaxPool}.

\subsection{\texttt{DetectionToTrackClassifier}}

Given the ROI features of a combination of detections and tracks $\mathbb{F}=(N \times M)$, the \texttt{DetectionToTrackClassifier} predicts a cost matrix of a match probability between detections and tracks $S=(N \times M)$. \texttt{DetectionToTrackClassifier} is simply a couple of layers of element-wise multilayer perceptrons (MLP) on $\mathbb{F}$ to predict match probability logits for logistic regression.

During training, we are given ground-truth tracks $\mathcal{T}_{\text{GT}}=\{\{\mathcal{T}^t_{GTi}\} \forall t \in [1, T]\} \forall i \in [1, K]$, where $K$ is the number of ground-truth tracks. Using $\mathcal{T}_{\text{GT}}$, we generate a set of positive and negative matching tracks with a combination of tracks at the current frame ($t=1$) and previous frames ($t\in[2,T]$). $$\{\mathcal{T}^1_{GTi}, \mathcal{T}^2_{GTj}, \mathcal{T}^3_{GTj}, \ldots, \mathcal{T}^T_{GTj}\} \forall i \in [1, K]\ \forall j \in [1, K]$$ $\mathcal{T}^1_{GTi}$ resembles detection $\mathcal{B}_i$ and $\{\mathcal{T}^2_{GTj}, \mathcal{T}^3_{GTj}, \ldots, \mathcal{T}^T_{GTj}\}$ resembles track $\mathcal{T}_j$ during evaluation. Then, we consider the matching tracks $(i=j)$ as positive and all else $(i\neq j)$ as negative matching tracks.

However, two issues arise with this approach. First, the positive-to-negative ratio $1:(K - 1)$ is heavily biased. Second, most of the negative samples are too easy, e.g. matching a pedestrian on one end of the map to the car on the other end of the map. To mitigate this issue, we have two fixes. First, we extract hard negative samples using a heuristic filter $f$. We define heuristic filter $f(\mathcal{T}^1_{GTi},\{\mathcal{T}^2_{GTj}, \mathcal{T}^3_{GTj}, \ldots, \mathcal{T}^T_{GTj}\})$ as follows: 1. $c(\mathcal{T}_{GTi})=c(\mathcal{T}_{GTj})$ where $c$ is a semantic class of an object 2. $dist(\mathcal{T}^1_{GTi}, \mathcal{T}^1_{GTj})<G$, where $dist$ is center-to-center 2D Euclidean BEV distance and $G$ is a hyperparameter gating distance. Any tracks that do not satisfy $f$ are filtered out to make the negative samples of the training data more meaningful. Second, we use binary focal loss~\cite{lin2017focal}, where $\hat{p}$ is the predicted match probability, $\alpha$ is a post-filter positive-to-negative ratio, and $\gamma=2$. $$L_{\text{track}}=-1[i=j]\alpha(1-\hat{p})^\gamma\log(\hat{p})-1[i\neq j]\hat{p}^\gamma\log(1-\hat{p})$$

The final training loss of \networkname is as follows, where $\lambda_{\text{track}}$ is a loss balancing term. $$L=L_{\text{det}} + \lambda_{\text{track}} L_{\text{track}}$$

\begin{table*}[t]
\centering
\begin{tabular}{c|ccccccccccccccccc|}
 Method & AMOTA↑ & MOTA↑ & RECALL↑ & FP↓ & FN↓ & IDS↓ & FRAG↓ \\
 \hline
 StanfordIPRL-TRI~\cite{chiu2020probabilistic} & 0.550 & 0.459 & 0.600 & 17533 & 33216 & 950 & 776 \\
 CenterPoint~\cite{centerpoint}$^*$ & 0.638 & 0.537 & 0.675 & 18612 & 22928 & 760 & 529 \\
 OGR3MOT~\cite{zaech2022learnable}$^*$ & 0.656 & 0.554 & 0.692 & 17877 & 24013 & 288 & 371 \\
 Belief Propagation~\cite{meyer2018message}$^*$ & 0.666 & 0.571 & 0.684 & 16884 & 22381 & \textbf{182} & \textit{245} \\
 SimpleTrack~\cite{pang2021simpletrack}$^*$ & 0.668 & 0.566 & 0.703 & 17514 & 23451 & 575 & 591 \\
 ImmortalTracker~\cite{wang2021immortal}$^*$ & 0.677 & 0.572 & 0.714 & 18012 & 21661 & 320 & 477 \\
 GNN-PMB~\cite{liu2022gnn}$^*$ & 0.678 & 0.563 & 0.696 & \textit{17071} & \textit{21521} & 770 & 431 \\
 NEBP~\cite{liang2022neural}$^*$ & 0.683 & \textbf{0.584} & 0.705 & \textbf{16773} & 21971 & \textit{227} & 299 \\
 TransFusion-L~\cite{bai2021pointdsc} & \textit{0.686} & 0.571 & \textit{0.731} & 17851 & 23437 & 893 & 626 \\
 \trackername (ours)$^*$  & \textbf{0.698} & \textit{0.578} & \textbf{0.757} & 19340 & \textbf{21220} & 325 & \textbf{217} \\
\end{tabular}
\caption{Tracking result on Nuscenes test set. Our proposed method achieved state-of-the-art performance on AMOTA metric. \trackername has the highest recall and lowest false negative, meaning that our 4D object detection trained jointly with online tracking recovers as many tracks as possible. \trackername also has the lowest track fragmentation and low ID switches demonstrating that our network successfully learns implicit motion models to assign detections to tracks.}
\label{table:nuscenes_test}
\end{table*}

\subsection{Track match and track confidence update}

\trackername updates track $\mathcal{T}$ given the output of \networkname detection bounding boxes $\mathcal{B}$ and detection-to-track match probability $\mathcal{S}$. Prior to assigning detections to tracks, the tracker filters out obvious mismatches using the aforementioned heuristic filter $f(\mathcal{B}_i, \mathcal{T}_j)$.

Additionally, in order to enforce geometric relations to the matching scheme, we also define distance ratio $\mathcal{D}=(N \times M)$ as $\mathcal{D}_{i\times j}=dist(\mathcal{B}_{i}, \mathcal{T}^2_{j} + {vel}_j^2)/G$. Since any tracks with $dist(\cdot, \cdot)>G$ are filtered out, the distance ratio is within [0, 1], the smaller the better. Then, we Hungarian match detections $\mathcal{B}$ to tracks $\mathcal{T}$ to minimize the cost matrix $\lambda_{D}\mathcal{D} - (1 - \lambda_{D}) \mathcal{S}$, where $\lambda_{D}$ is a balancing hyperparameter. Any unmatched detections are initialized as a new track. Any tracks unmatched for longer than $T$ are considered as dead tracks.

Lastly, we estimate the confidence for each track. Track confidence scores determine the reliability of a track which is crucial for its applications. Thus, the tracking evaluation metric AMOTA~\cite{amota} takes track confidence into account. The most commonly defined track confidence score is an average of detection confidence scores $s_{\text{det}}$ in a track. This is a reasonable choice since a track is composed of detections. However, it lacks information on the confidence of associations among the detections. Therefore, we complement the missing information by defining track score as a track average of $(1 - \lambda_{S})s_{\text{det}i} + \lambda_{S}\mathcal{S}_{i \times j}$, a linear combination of a detection score and its detection-to-track match probability score. In our experiment, we will quantitatively confirm that our score incorporating the detection-to-track match probability improves AMOTA.

\section{Implementation details}

Our implementation is based on an open-source implementation of CenterPoint~\cite{centerpoint} by OpenMMLab. Our encoder architecture follows that of SECOND~\cite{yan2018second} with some changes to process the 4D data, as described in Method section 1. The last layer of the encoder is composed of three deformable convolution~\cite{dai2017deformable} to learn different features for classification, bounding box regression, and instance prediction. The classification and bounding box regression features are consumed by the first stage CenterPoint head, as discussed in Method section 2. The instance prediction features are consumed by the second stage tracker in Method section 3.

For hyperparameters, the gating distance $G$, following the tracker scheme of CenterPoint, is based on per-class 99 percentile error of velocity prediction. More precisely, 4m for cars and trucks, 5.5m for buses, 3m for trailers, 1m for pedestrians, 13m for motorcycles, and 3m for bicycles. Focal loss term $\alpha$ is a post-filtering per-class positive-to-negative ratio, 4 for cars and pedestrians, and 2 for all else. The detection and track loss balancing term $\lambda_{\text{track}}=1$. The track assignment cost matrix balancing parameter $\lambda_D=0.5$ and track confidence score balancing term $\lambda_S=0.2$, as shown in our ablation study.

For Nuscenes dataset, we crop point clouds at range $W = [-54, 54], H = [-54, 54], D= [-5, 3]$ and voxelize it at size $(v_W, v_H, v_D) = (0.075, 0.075, 0.2)$ for X, Y, and Z axis respectively. The temporal axis size $T=3$. Each $t$ is 0.5 seconds following the 2HZ sampling rate of tracking key sequences. Each $t$ is composed of a concatenation of 10 samples of high-frequency lidar point clouds. We train our network for 30 epochs with AdamW~\cite{loshchilov2018fixing} optimizer of learning rate of 1e-4 and weight decay 1e-2 with a one-cycle learning rate scheduler. We trained our model with a batch size of 16 on 8 TITAN RTX GPUs.
\section{Experiments}

\subsection{Dataset}

The Nuscenes dataset~\cite{caesar2020nuscenes} is a large-scale autonomous driving dataset. It comprises 1000 driving scenes of 20 seconds length. Each scene is labeled with 3D bounding boxes of 7 object classes at 2Hz to facilitate multi-object tracking algorithms. The dataset provides multiple modalities of sensors, while we only use lidar point clouds as input to our system. Following the tracking challenge protocol, our temporal window at $t$ is roughly 0.5 seconds.

\subsection{Metrics}

Following the Nuscenes tracking challenge, we use the following metrics to evaluate our tracker: AMOTA~\cite{amota}, MOTA~\cite{mota}, RECALL, FP (False Positive), FN (False Negative), IDS (track ID switches), and FRAG (track FRAGmentation). MOTA has long been a standard evaluation metric for object tracking, measuring the common track association errors such as FP, FN, and IDS. AMOTA takes track confidence scores into account, being an integral of MOTA over recall scores. AMOTA is the primary metric of the Nuscenes tracking challenge.

\subsection{Baseline model}

CenterPoint~\cite{centerpoint} is our primary baseline 3D object detection and online tracking model. Centerpoint built a simple yet powerful 3D online tracking baseline by matching tracks with the closest object detection backprojecting the predicted velocity to the track. Our method is agnostic to the choice of a 3D object detector. We chose CenterPoint since most of the previous works in the Nuscenes tracking challenge take the detection result of CenterPoint as an input to the tracker.

\subsection{Results}

We tested our method on the test set of the Nuscenes tracking challenge. As shown in Tabel~\ref{table:nuscenes_test}, our method achieved the state-of-the-art AMOTA of all published submissions based only on lidar point clouds. \trackername has notably high recall and lowest false negative among all methods. This result indicates that our method is superb at recovering ground-truth tracks with limited observations, owing to the 4D encoder's temporal reasoning and multi-task training of detection and tracking. \trackername also has the lowest track fragmentation and low ID switches demonstrating that the network's implicit motion model successfully connects the tracks. Overall, our proposed framework quantitatively outperformed previous works in various metrics.

It is also worth noting that the performance of multi-object tracking is bound to that of object detection. \trackername uses CenterPoint as a object detection head, with other CenterPoint-based trackers marked as ${}^*$ in Table~\ref{table:nuscenes_test}. Our network is modular, and the latest development of 3D object detection can easily be incorporated by updating the detection head.

\begin{figure}[t]
    \centering
    \begin{tabular}{cccc}
        &$t=1$&$t=2$&$t=3$ \\
        \rotatebox{90}{CenterPoint} &
        \includegraphics[width=0.12\textwidth]{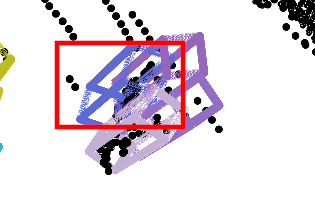} &
        \includegraphics[width=0.12\textwidth]{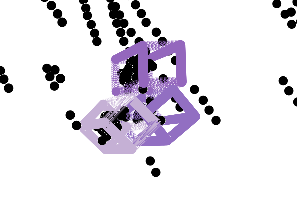} &
        \includegraphics[width=0.12\textwidth]{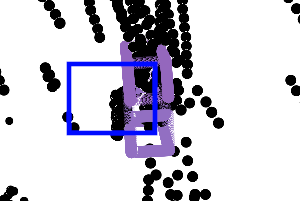} \\
        \rotatebox{90}{Ours} &
        \includegraphics[width=0.12\textwidth]{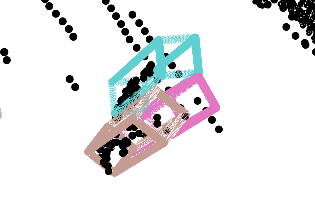} &
        \includegraphics[width=0.12\textwidth]{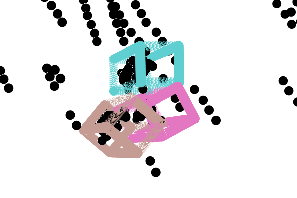} &
        \includegraphics[width=0.12\textwidth]{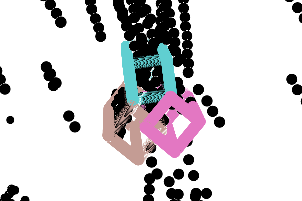} \\
    \end{tabular}
    \begin{tabular}{cccc}
        \rotatebox{90}{CenterPoint} &
        \includegraphics[width=0.12\textwidth]{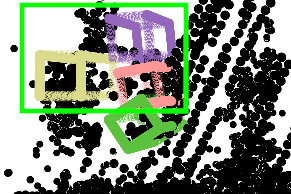} &
        \includegraphics[width=0.12\textwidth]{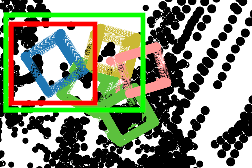} &
        \includegraphics[width=0.12\textwidth]{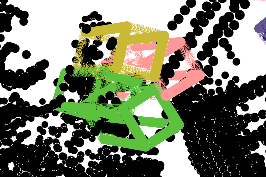} \\
        \rotatebox{90}{Ours} &
        \includegraphics[width=0.12\textwidth]{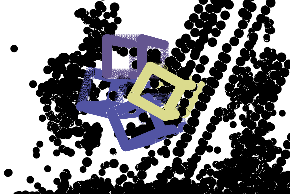} &
        \includegraphics[width=0.12\textwidth]{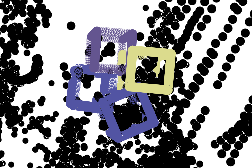} &
        \includegraphics[width=0.12\textwidth]{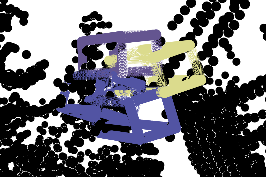} \\
    \end{tabular}
    \begin{tabular}{cccc}
        \rotatebox{90}{CenterPoint} &
        \includegraphics[width=0.12\textwidth]{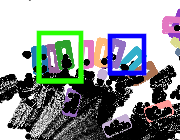} &
        \includegraphics[width=0.12\textwidth]{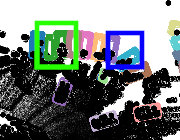} &
        \includegraphics[width=0.12\textwidth]{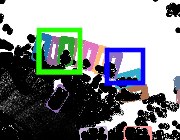} \\
        \rotatebox{90}{Ours} &
        \includegraphics[width=0.12\textwidth]{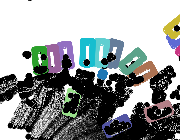} &
        \includegraphics[width=0.12\textwidth]{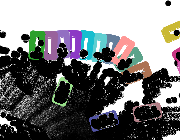} &
        \includegraphics[width=0.12\textwidth]{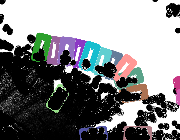} \\
    \end{tabular}
    \caption{Qualitative results comparing tracking result of CenterPoint with \trackername. Our method predicts consistent tracklets of pedestrians, being more robust to \textcolor{red}{false positives}, \textcolor{blue}{false negatives}, and \textcolor{green}{ID switches.}}
    \label{fig:qualitative2}
\end{figure}

In Figure~\ref{fig:qualitative2}, we visualized a qualitative result of tracks on the validation set of the Nuscenes dataset, comparing CenterPoint with \trackername. First, CenterPoint suffers from detection errors such as false positives and false negatives. In contrast, our method predicts persistent detections benefiting from the temporal reasoning of the 4D encoder. This result aligns with the quantitative result that our method achieved the best recall among all methods. Furthermore, CenterPoint suffers from ID switches, whereas our method robustly tracks the target, demonstrating that our implicit motion model successfully assigns detections to tracks. This result explains why our method achieved the lowest track fragmentation among all methods and less than half ID switches compared to CenterPoint.

\subsection{Ablation Study}
\begin{table}[t]
\centering
\begin{tabular}{c|cc}
 Method & mAP↑ & NDS↑ \\
 \hline
 CenterPoint & 59.6 & 66.8 \\
 \hline
 CenterPoint + 4DEncoder  & 61.9 & 69.3 \\
 CenterPoint + 4DEncoder + \textit{track}  & 62.4 & 69.5 \\
\end{tabular}
\caption{
    Detection result comparison on Nuscenes detection validation split between CenterPoint~\cite{centerpoint} and \networkname. 4DEncoder refers to our network using a 4D sparse encoder. \textit{track} refers to a detector jointly trained with a tracking loss.
}
\label{table:detection}
\end{table}

\begin{table}[t]
\centering
\begin{tabular}{c|c}
 Method & AMOTA↑ \\
 \hline
 CenterPoint & 66.9 \\
 \hline
 CenterPoint + 4DEncoder  & 67.4 \\
 CenterPoint + 4DEncoder + \textit{track}   & 67.8 \\
 CenterPoint + 4DEncoder + \textit{track} + \textit{score}  & 69.1 \\
 CenterPoint + 4DEncoder + \textit{track} + \textit{conf}  & 68.5 \\
 CenterPoint + 4DEncoder + \textit{track} + \textit{score} + \textit{conf}  & 70.3 \\
\end{tabular}
\caption{
    Multi-object tracking ablation study on Nuscenes tracking validation split. 4DEncoder refers to our network using a 4D sparse encoder. \textit{track} refers to a detector jointly trained with a tracking loss. \textit{score} refers to the track assignment based on detection-to-track score. \textit{conf} refers to the detection-to-track score-based track confidence.
}
\label{table:ablation}
\end{table}

We demonstrate in large-scale experiments that the overall performance gain of our method is due to four factors: 1. The 4D encoder improves detection by adding temporal reasoning. 2. Multi-task learning of object detection and online tracking enhances each other. 3. The network learns the implicit motion model to improve track assignment 4. The track confidence score incorporating the detection-to-track score is informative. In this section, we thoroughly analyze the performance gain of each component.

First, the 4D encoder of \networkname improves the object detection performance, subsequently improving the tracking performance. In Table~\ref{table:detection} and~\ref{table:ablation}, we compare the detection and tracking performance of CenterPoint with the original 3D encoder and our 4D encoder. \networkname outperforms CenterPoint in object detection owing to the extended horizon of temporal reasoning through the 4D sparse encoder. Subsequently, the tracking result of the 4D detector global nearest neighbor match outperforms CenterPoint.

Second, the multi-task learning of detection and tracking jointly enhances the performance of each other. In Table~\ref{table:detection} and~\ref{table:ablation}, we compare the detection and tracking performance of the 4D encoder CenterPoint, trained with and without the tracking loss ($\lambda_{\text{track}}=\{1, 0\}$). A detector and tracker trained with the tracking loss (\textit{track}) outperforms the bare detector. The improved detection from the 4D encoder and multi-task learning explains the quantitative result that our method has the highest recall and lowest false negative among all methods.

\begin{figure}[ht]
    \centering
    \begin{tabular}{cc}
        \includegraphics[width=0.45\textwidth]{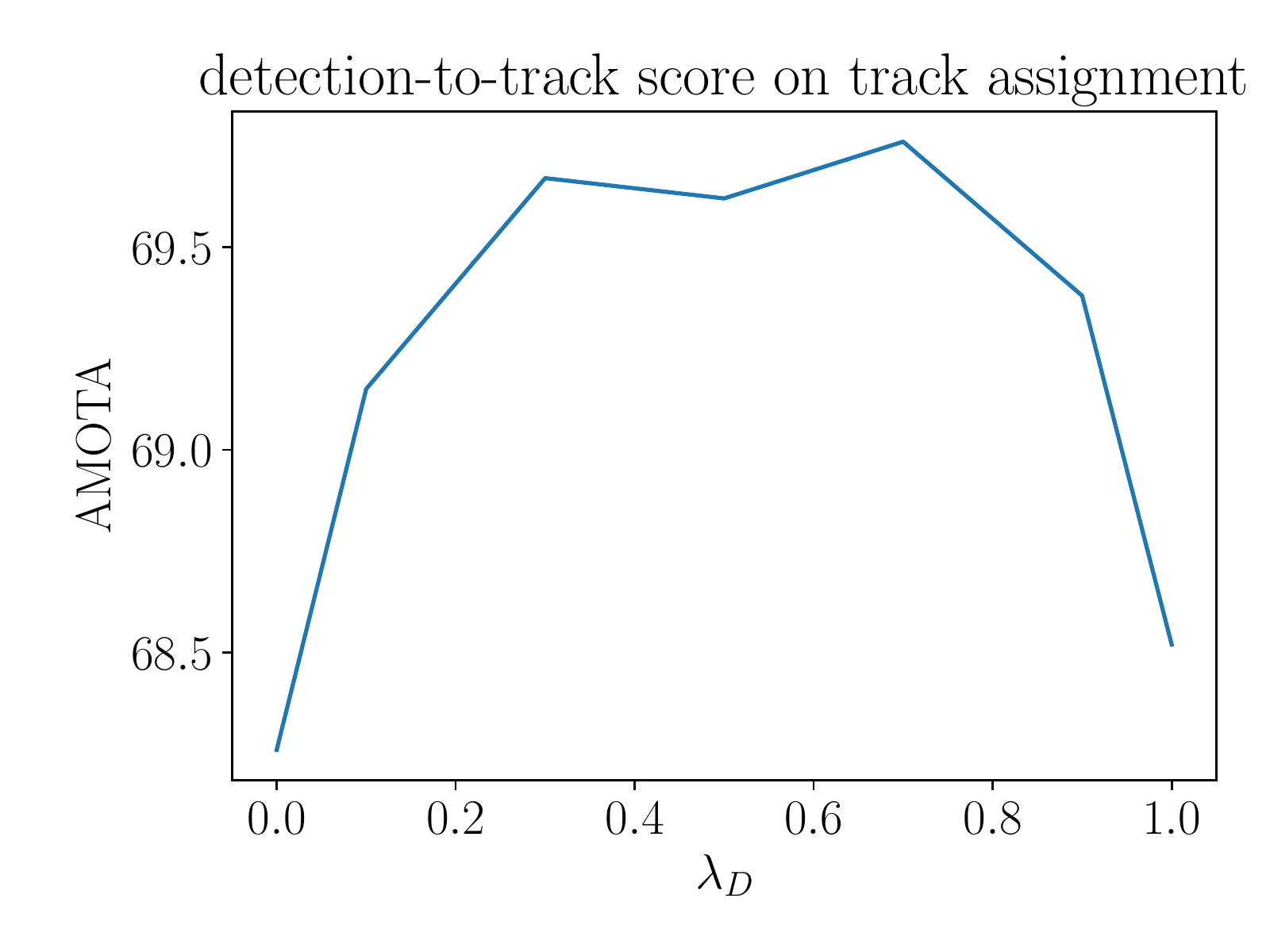} \\
        \includegraphics[width=0.45\textwidth]{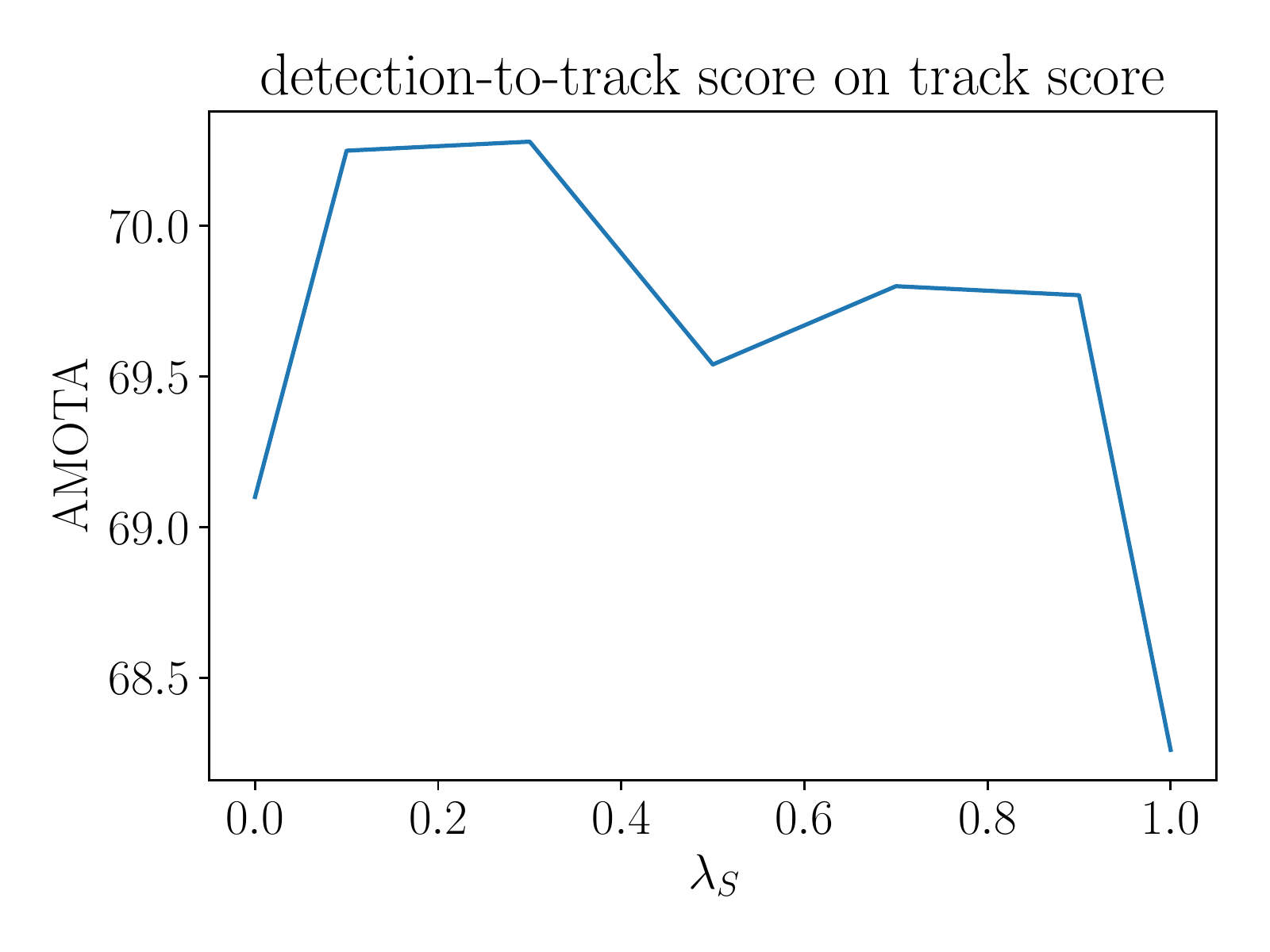}
    \end{tabular}
    \caption{
        Ablation study of the impact of a detection-to-track score on track assignment and confidence score. $\lambda_D=1$ is a global nearest neighbor matching similar to the tracking algorithm of CenterPoint, whereas $\lambda_D=0$ is a matching solely based on our proposed detection-to-track score. $\lambda_S=0$ is a tracking confidence score only based on detection confidence, whereas $\lambda_S=1$ is a tracking confidence score only based on the detection-to-track score.}
    \label{fig:ablation}
\end{figure}

\begin{figure}[t]
    \centering
    \includegraphics[width=0.45\textwidth]{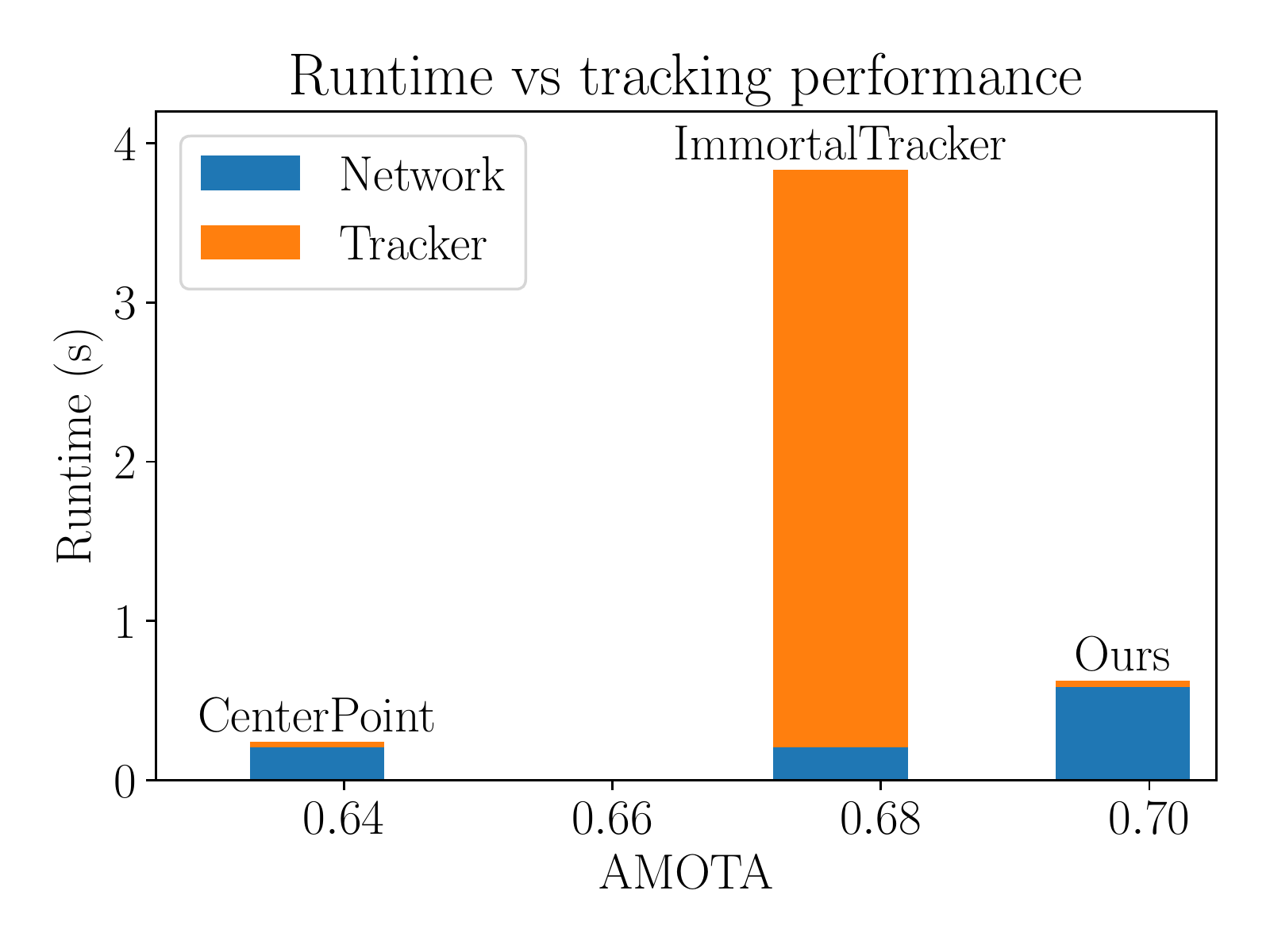}
    \caption{
    Runtime analysis and comparison with other works. The x-axis is tracking performance, and the y-axis is the tracker's runtime, separated into the runtime of the neural networks and the tracker. The runtime bar chart of each tracker is plotted on the x-axis of the corresponding tracking performance.
    }
    \label{fig:runtime}
\end{figure}

Third, the cost matrix incorporating the detection-to-track match probability score improves the quality of the track assignment. As shown in Table~\ref{table:ablation}, the advanced track assignment with the learned score, denoted as \textit{cost}, gives a persistent quantitative improvement in the quality of tracks. This result indicates that our network learns an implicit motion model that improves track assignment, which explains the lowest track fragmentation and low ID switches of the quantitative results.

Fourth, the detection-to-track score prediction by \networkname improves the quality of the track confidence. As shown in Table~\ref{table:ablation}, the track confidence incorporating the matching score, denoted as \textit{conf}, gives a persistent quantitative improvement in the AMOTA metric, which is a tracking metric that takes track confidence score into account.

\subsection{Hyperparameter Search}

In Figure~\ref{fig:ablation}, we measure the performance of our model with varying hyperparameters. First, the network predicted matching score $\mathcal{S}$ and distance ratio $\mathcal{D}$ are both effective measures for track assignment. Nevertheless, a linear combination of $\mathcal{D}$ and $\mathcal{S}$ gives a noticeable improvement in tracking performance, hinting that there are complimentary information between the two.

Second, the combination of the network predicted matching score $\mathcal{S}$ and the detection confidence $s_{\text{det}}$ results in more robust track confidence. The fact that a track is composed of detections and their temporal associations explains why track confidence should be composed of both detection confidence $s_{\text{det}}$ and association score $\mathcal{S}$.

\subsection{Runtime Analysis}

In Figure~\ref{fig:runtime}, we analyze and compare the runtime of \trackername with CenterPoint~\cite{centerpoint} and ImmortalTracker~\cite{wang2021immortal}. The runtime is measured on a single-core AMD Ryzen 7 1800X CPU with a single TITAN RTX GPU. The primary concern of the runtime of our method is that we use a 4D encoder, which requires at least $\times T$ more computation compared to its 3D counterpart. Not surprisingly, the network runtime of our method is roughly $T=3$ times slower than CenterPoint.

However, \trackername does not require complicated hand-designed filters to achieve high performance. CenterPoint relatively does not perform well in tracking since it uses a bare minimal motion model. ImmortalTracker, based on CenterPoint, requires a much higher runtime in the tracker to compute 3D Kalman Filter with linear motion model and 3D GIoU~\cite{rezatofighi2019generalized} for a reliable matching. In contrast, \trackername relies on the network's implicit motion model, taking a simple linear assignment of the detection-to-track score, which does not require much runtime in the tracker. As a result, \trackername is $\times 6$ faster than ImmortalTracker while achieving the state-of-the-art tracking performance.

\section{Conclusion}
We propose \trackername, a sparse spatio-temporal R-CNN for joint object detection and tracking. Our network processes the 4D point cloud using a 4D sparse encoder to improve detection with temporal reasoning. Our proposed \texttt{TrackAlign} enables two-stage joint detection and tracking for improved multi-task learning. Furthermore, our network learns an implicit motion model, which advances track assignment and confidence score. As a result, \trackername reached the state-of-the-art performance on the Nuscenes tracking challenge, achieving the highest recall and lowest false negative, validated in our qualitative result and ablation study.

\section{Acknowledgement}
Toyota Research Institute provided funds to support this work.

\bibliography{refs}

\end{document}